\documentclass[11pt,a4paper]{article}
\usepackage[hyperref]{naaclhlt2019}
\usepackage{booktabs}
\usepackage{times}
\usepackage{latexsym}
\usepackage{nameref}
\usepackage{varioref}
\usepackage{hyperref}
\usepackage{graphicx}
\usepackage{caption}
\usepackage{subcaption}
\usepackage{url}
\usepackage{bm}
\usepackage{amsmath, amssymb}
\usepackage{algorithm, algorithmic}
\usepackage{cleveref}

\aclfinalcopy % Uncomment this line for the final submission

\title{A  dataset for resolving referring expressions in spoken dialogue via contextual query rewrites (CQR)}
\author{Michael Regan \\ {\tt reganman@unm.edu} \\University of New Mexico\\ Albuquerque, NM,USA \\\And
Pushpendre Rastogi \\ {\tt pushpendre@jhu.edu} \\Johns Hopkins University\\ Baltimore, MD,USA \\\And
Arpit Gupta \\ {\tt arpgup@amazon.com} \\Alexa AI \\Amazon.com, Inc., USA  \\\AND
Lambert Mathias\\ {\tt mathiasl@amazon.com} \\Alexa AI \\Amazon.com, Inc., USA }

\date{}

\begin{document}
\maketitle
\begin{abstract}

We present Contextual Query Rewrite (CQR) a dataset for multi-domain task-oriented spoken dialogue systems that is an extension of the Stanford dialog corpus ~\cite{Eric2017}.
While previous approaches have addressed the issue of diverse schemas by learning candidate transformations ~\citep{naik2018contextual}, we instead model the reference resolution task as a user query reformulation task, where the dialog state is serialized into a natural language query that can be executed by the downstream spoken language understanding system. In this paper, we describe our methodology for creating the query reformulation extension to the dialog corpus, and present an initial set of experiments to establish a baseline for the CQR task. We have released the corpus to the public\footnote{\url{https://github.com/alexa/alexa-dataset-contextual-query-rewrite}} 
to support further research in this area.

\end{abstract}

\section{Introduction}
\label{sec:intro}
Spoken language understanding (SLU) for dialogue systems is designed to correctly interpret a user utterance, a unit of communication given a specific context \cite{Grice1968}. For 
successful interpretation, the system may classify user intent (the action that the user would like the system to perform) and resolve slot values (particular attributes associated with 
intents). Task-oriented dialogues currently most often consist of a one-shot utterance said by the user for the digital assistant to perform a task (e.g., `What is the weather like?'). In 
some cases, however, a dialogue may involve multiple turns, each turn a sequence of user and system utterances. Reasoning over multi-turn dialogues, in which user and agent add 
information incrementally to specify the user's intent, is a challenge which only increases in difficulty when the agent needs to resolve referring expressions across turns, either 
explicit references (example - pronominal and nominal anaphora) or implicit ones (example - zero anaphora).  

In this paper, we approach the referring expression resolution task in multi-turn discourse as a query reformulation task - the utterance containing the referring expression is rewritten to 
contain all relevant slot values from the context, a process we call \textit{contextual query rewrite} (CQR). In order to be feasible, query reformulation must be able to leverage
multi-turn context, be intuitive, and learnable, principles we apply in the creation of our corpus extended via crowd-sourcing and then used to train an end-to-end dialogue system 
without a need for explicit state trackers. Our main contributions are:
\begin{enumerate}
\item We introduce a new task - contextual query rewrite - for resolving referring expressions without explicit need to track the state.
\item We release a publicly available corpus consisting of gold standard and crowd-sourced rewrites as an extension to an existing task-oriented dialogue corpus.
\end{enumerate}

\section{Problem}
\label{sec:approach}
\subsection{Motivation}
We motivate contextual query rewriting (CQR) with an example	 shown in Table~\ref{tbl:queryexample}. Typically, we would expect a digital assistant to understand the second user 
utterance (U2) as referring to traffic near the coffee shop, rather than defaulting to the user's current location, although here this is not explicitly stated. Our correct interpretation of this 
utterance is possible via an implicit reference to the location using \textbf{zero} anaphora (e.g., ``What's the address (of the coffee shop)?''), recognized as the predominant anaphora 
type observed cross-linguistically \cite{Givon2017}. The user could also refer to the same coffee shop using a \textbf{nominal} anaphoric reference (``that coffee shop'') , or a 
\textbf{locative} form (``there''), or as a \textbf{pronominal} form (e.g., ``it'').

\begin{table*}[h!]
\small
\begin{center}
\begin{tabular}{|c|c|c|c|c|}
\hline
speaker  & domain & utterance &  Resolved referenced slots(key=value) & Reformulated query\\
\hline\hline
U1 &  POI & any coffee shops nearby & poi\_type=coffee shop &\\
& & & distance=nearby &\\\hline
V1 & POI & found a coffeeworks 2 miles away & poi=coffeeworks &\\
 &  & & distance=2 miles &\\\hline
U2 & Traffic & how's the traffic & location=coffeeworks & how's the traffic to coffeeworks\\\hline
\end{tabular}
\end{center}
\caption{\label{tbl:queryexample} Multi-turn dialogue with implicit reference in (U2) to the slot (poi=coffeeworks). The turn U2 is handled by the Traffic domain NLU, with its own domain 
schema - the  slot (poi=coffeeworks) needs to be transformed to the slot (location=coffeeworks). Contextual query rewriting (CQR) would resolve the implicit referring expression by 
reformulating the query as shown for U2 in a schema agnostic way.}

\end{table*}

\begin{figure*}[ht!]
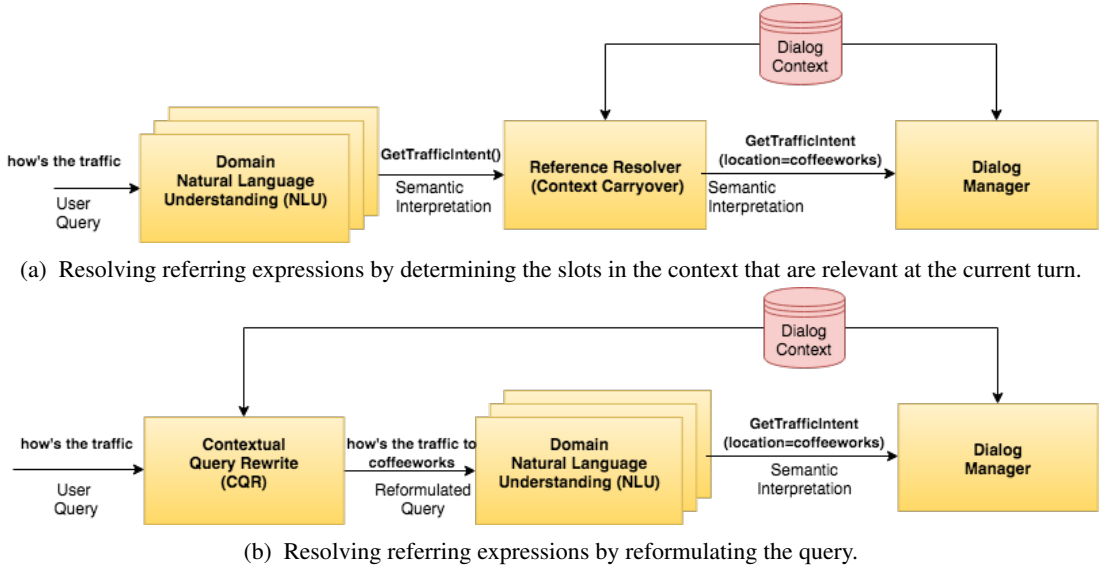

	\centering
	\begin{subfigure}{0.9\textwidth} % width of left subfigure
		\includegraphics[width=\textwidth]{context_carryover_architecture.png}
		\caption{\label{fig:cc} Resolving referring expressions by determining the slots in the context that are relevant at the current turn.} 
		 % subcaption
	\end{subfigure}
	\vspace{1em} % here you can insert horizontal or vertical space
	\begin{subfigure}{0.9\textwidth} % width of right subfigure
		\includegraphics[width=\textwidth]{cqr_architecture.png}
		\caption{\label{fig:cq} Resolving referring expressions by reformulating the query.} % subcaption
	\end{subfigure}
	\caption{Contextual Query Rewrite Architecture - Referring expression resolution happens before running the language understanding component, compared to the traditional 
approach where the reference resolution usually takes place after language understanding.} % caption for whole figure
\end{figure*}

More generally, the task of referring expression resolution can be solved as a carryover task, where the relevant slots from the context are carried over to the current turn~
\citep{naik2018contextual}. For our working example described in Table~\ref{tbl:queryexample}, the result of the carryover task shows up as additional slots associated with the utterance,  as show in in Figure~\ref{fig:cc}. A challenge is dealing with domain specific schemas requiring accurate transformations even for emergent slots where there is very little data available  to train these mappings correctly. Another solution is to make the natural language understanding system contextually aware~\cite{gupta2018efficient}. However, updating the domain-specific NLU sub-systems is more complex, as it requires re-training the production sub-systems, often a time-consuming and laborious task. Moreover, this approach does not work for systems that model the meaning using frameworks other than intents and slots.

In this paper, we propose query reformulation, where we take an otherwise ambiguous utterance such as ``how's the traffic'' in Table \ref{tbl:queryexample} and add the relevant slot 
values from the context, here the name of the place (``coffeeworks''), to make a reformulated query: ``how's the traffic to coffeeworks?'', as shown in Figure~\ref{fig:cq}. We call this 
approach to resolving referring expressions {\it contextual query rewrite} (CQR). The main advantage of this approach is that it does not require updating the domain specific NLUs, 
and takes advantage of the fact that these systems are optimized for single shot performance. Resolving referring expressions is now equivalent to generating a single shot natural 
language query, thereby making this process invariant to the meaning representation and domain specific schema changes.

\subsection{CQR Task Formulation}
\label{ssec:cqr}
We can now formally define the CQR task.  We define a sequence of $D$ dialogue turns $\{\bm{u}_{t-D+1}, \dots, \bm{u}_{t-2}, \bm{u}_{t-1}\}$; the current user utterance $\bm{u}_{t}\}
$\footnote{For simplicity we assume a turn taking model - a user turn and system turn alternate.}, where $\bm{u} \in \mathcal{W}$ is  a sequence of tokens $\{w_{i}\}_{i=1}^{N}$. 
Associated with the $D$ dialogue turns and the current turn is a set of slot values $\bm{v}=\{v_1, v_2, \cdots, v_N\}$. The CQR task is to learn a mapping
\begin{equation*}
\mathcal{F}(\bm{u}_t \cup D) \rightarrow \bm{u'}_t
\end{equation*}

\noindent where the reformulated user turn $\bm{u'}_t$ contains tokens copied either from the vocabulary $\mathcal{W}$ or from a subset of relevant slot values from $\bm{v}$. The challenge is to learn a reformulated user utterance that has implicitly selected the subset of slots that are relevant at turn $t$, while retaining the semantics associated with turn $t$.

\section{Related work}
\label{sec:relatedwork}

\textbf{Dialogue corpora}: There are various dialogue corpora, and collection methodologies.~\cite{Weston2015} have the objective of improving algorithms for text understanding by 
modularizing each reasoning task; two of their tasks involve coreference; however, these appear to involve the resolution of pronominal forms exclusively, forms which in on our data 
represent only a small fraction of all anaphoric references.~\cite{Bordes2017} released a corpus of 18,000 synthetic dialogues for a single domain (restaurant reservations), however, 
these do not reflect real user behavior. Human efforts may also be directed in a Wizard-of-Oz schema using the interactions of crowd-sourced workers to develop corpora. For example, \cite{Wen2017} create a corpus of approx. six hundred eighty dialogues for a single domain (restaurants), and like them we also set out to avoid handcrafting and labeled datasets by representing slot-value pairs explicitly.  \cite{Eric2017} use the same approach to create $3k$ dialogues for three domains (weather, navigation, and calendar scheduling), a corpus particularly rich with anaphoric references.
 
\textbf{State tracking} Dialogue state tracking (DST) is considered to be a higher-level module as it has to combine information from previous user utterances and system responses with the current utterance to infer full meaning. Many deep-learning based methods have recently been proposed for DST such as a neural belief tracker~\cite{mrkvsic2017neural}, and a self-attentive dialogue state tracker~\cite{zhong2018global} which are suitable for small-scale domain-specific dialogue systems; as well as more scalable approaches such as~\cite{rastogi2017scalable,xu2018e2e} that solve the problem of an infinite number of slot values and~\cite{naik2018contextual} who additionally solve the problem of large number of disparate schemas in each domain. Unfortunately, all of the above approaches fail to address the problem that as the number of domain-specific chatbots on a dialogue platform grows larger, the DST module becomes increasingly complex as it tries to handle the interactions between different chatbots and their different schemas.

\textbf{Text Generation} Seq2Seq models with attention~\cite{sutskever2014sequence,bahdanau2014neural} have seen rapid adoption in automatic summarisation~
\cite{see2017gttp,rush2015}.~\cite{mem2seq2018,eric2017key} propose end-to-end generative approaches where a copy mechanism is used to copy entities from a knowledge base when generating the response. Closest to this work, is the copy mechanism based user query reformulation for search-based systems~\cite{dehghani2017learning}. Exploring black-box methods like query re-writing allow us to benefit from the progress made in these fields and apply them to state tracking and reference resolution tasks in dialogue.

\section{Contextual Query Rewrite Dataset}
\label{sec:corpus}

\subsection{Query reformulation methodology}
\label{sec:principles}

To build a corpus of query reformulations, we begin first with the principles that guided our decision-making process. 
\begin{enumerate}\setlength\itemsep{0.1em}
\item {\bf Multi-turn}: We expect human-computer interaction will soon more often involve multiple turns (where each turn consists of a single user and agent utterance pair); anaphoric 
references are likely to occur more in multi-turn discourse; and, cross-linguistically, anaphor is a standard linguistic strategy for referring to the same entity and increase discourse 
cohesiveness \cite{Hobbs1979} signaling what new versus known information is; within-sentence anaphoric references fall outside the scope of the present research framework;
\item {\bf Intuitive}: Deciding which slot values are relevant given a particular dialogue history should be intuitive, assessed as the agreement among individuals; 
\item {\bf Interpretable}: Evaluating the output of a model should be straightforward, i.e., given the guidelines for query reformulation (below), an analyst will be able to assess quickly 
performance;
\item {\bf Learnable}: An end-to-end dialogue system should be able to resolve anaphoric references to increase user satisfaction; the extent to which a model learns to identify which 
slot values are relevant can be examined, explored in Section \ref{sec:experiments}.
\end{enumerate}

The guidelines for the task of query reformulation are summarized here:

\begin{enumerate}\setlength\itemsep{0.1em}
\item Identify the utterance which most closely matches the user's intent or request; we call this the \textit{basis utterance};
\item Reformulate the basis utterance to be a one-shot utterance, making the user's intent unambiguous by including all relevant slot values from the previous context; determining what is 
`relevant', however, is dependent on how much we assume a model may automatically be able to infer given a specific utterance;
\item When a place is not referred to directly by poi\_type (e.g., ``I want some pizza''), the poi\_type is assumed to be inferable, e.g., as ``pizza restaurant''.
\item Some cases are not subject to reformulation, e.g., confirmation of the agent's decision (e.g., Agent: ``Would you like directions?'', User: ``Yes please''), or when giving thanks or 
otherwise signaling the end of the dialogue;
\item Slot values from the context replace an anaphoric reference (whether nominal, pronominal, or zero) in the basis utterance;
\item Only certain types of anaphora need to be attended to, specifically references to slot values from the given domain; we ignore non-slot values (e.g., ``\underline{We} want coffee'' 
does not need to be resolved to ``you and I''), as well as anaphoric references to propositions or events (e.g., ``\underline{That} sounds good'');
\item When multiple values for a specific slot are available from context or in the current utterance, only use the most specific slots values, e.g., ``Take me via the fastest route'', 
specifying route conditions precludes the need to specify traffic information further, e.g., ``avoid all heavy traffic'';
\item For utterances with multiple anaphora (e.g., ``Give me the address and let's go there''), resolve both references: e.g., ``Give me the address of the coffee shop Coupa and let's go 
to the coffee shop Coupa''); this is not strictly enforced;
\item Intent may need to be carried over in addition to slots, e.g., ``How about another coffee shop?'' is reformulated as ``Give me directions to another coffee shop...''; this is rare;
\end{enumerate}

\subsection{Corpus selection and first modifications}

\begin{table*}[h]
\centering
\begin{tabular}{lrrrrr}
\toprule
{} &  \#reformulations &    \#zero &  \#pronominal &  \#locative &  \#nominal \\
\midrule
dev   &       206 &   143 &        31 &      34 &     21  \\
test  &       214 &   155 &        21 &      43 &     20  \\
train &      1867 &  1138 &       393 &     162 &    143 \\\hline
Total   &      2287 &  1436 &       445 &     239 &    184  \\
\bottomrule
\end{tabular}
\caption{\label{tab:anaphora} Distribution of anaphora (zero, pronominal, locative, nominal) in modified corpus.}
\end{table*}

With the principles above (multi-turn dialogue with cross-sentential anaphora), we selected a publicly available corpus \cite{Eric2017}, a corpus composed of approximately three thousand dialogues over three domains (weather, scheduling, and navigation). For additional statistics regarding the original corpus, we refer the reader to \cite{Eric2017}. Applying the guidelines above to the first task of query reformulation, we arrived at a modified corpus to begin our study and later experiments, noticing primarily the anaphora types described in Section \ref{sec:intro}. 
In the released corpus, we include flags for each anaphora type\footnote{We also include flags for other interesting linguistic forms, incl. \textbf{either} (as in, ``Let's go to either...'') and 
\textbf{besides}, to exclude an option (e.g., ``Let's go to another coffee shop besides...'')}. 

\begin{figure}
\centering
\includegraphics[scale=0.55]{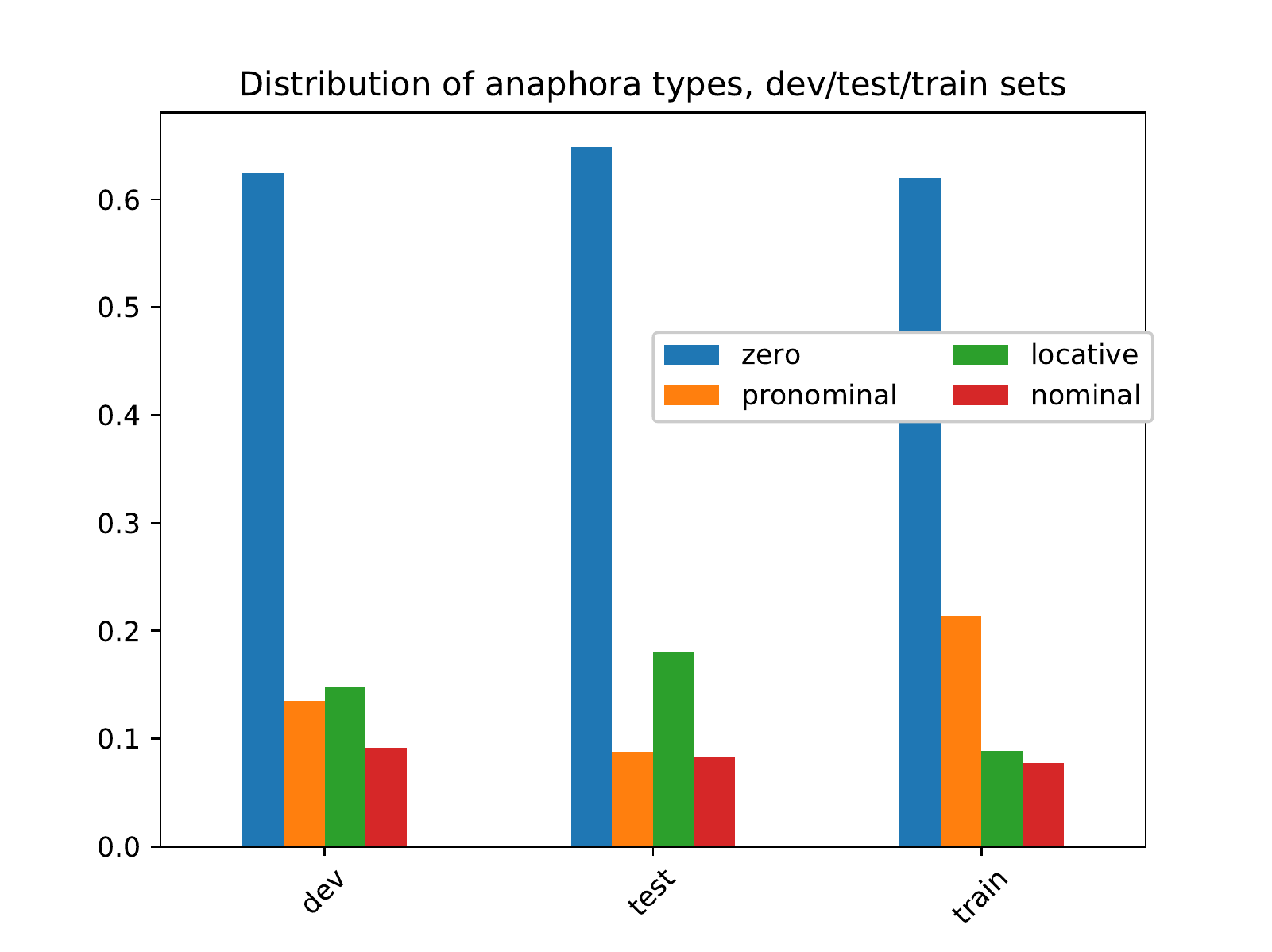}
\caption{\label{fig:anaphora}Distribution of anaphora (dev/test/train).}
\end{figure}

Figure \ref{fig:anaphora} shows the final distribution of anaphora types in the modified corpus, with total counts for anaphora types shown in Table \ref{tab:anaphora}. As an initial 
estimate of how much more information the gold reformulations contain, we count slot values in the basis utterances before and after reformulation, presented in Table 
\ref{tab:slotcounts}.

% Unnecessary table

%\begin{table}
%\small
%\centering
%\begin{tabular}{lrr}
%\toprule
%{} & either &  besides \\
%\midrule
%dev   &    11 &      4 \\
%test  &      6 &      4  \\
%train &     55 &     28  \\
%Total   &       72 &     36    \\
%\bottomrule
%\end{tabular}
%\caption{\label{tab:either}Distribution of two additional linguistic phenomena (`either' and `besides') in modified corpus.}
%\end{table}

\begin{table}[!h]
%\small
\centering
\begin{tabular}{lrrr}
\toprule
{} & (before) &  (after) & \# slots (after) \\
\midrule
dev   & 7.6 &           17.1   & 2.8\\
test  &  7.3 &           16.8   & 2.9 \\
train &  7.4 &           18.2   & 3.1\\
\bottomrule
\end{tabular}
\caption{\label{tab:slotcounts}Average number of tokens in each utterance before and after reformulation with the average number of slots in the reformulations.}
\end{table}

\subsection{Crowdsourcing reformulations}
\label{ssec:extension}

Upon completion of the first step of corpus modification, we then manually identified the basis utterance with relevant slot values for each dialogue and produced a set of gold reformulations. For scalability, we then extended the corpus by gathering five crowd-sourced reformulations for each dialogue, referred to here as \textit{rewrites} to be distinguished from the \textit{gold reformulations} we composed ourselves. 

We presented crowd-sourced workers with dialogues with highlighted basis utterance and the list of slot values that we judged relevant. Workers were encouraged to use their own strategies to rewrite the basis utterance in order to increase linguistic variation. Examples, instructions, and feedback to the workers helped ensure that semantic similarity (regarding
which slot values to include) was as high as possible. This process of collecting rewrites took two weeks, with other descriptions shown in Table \ref{tab:mturk}.

\begin{table}[!h]
%\small
\begin{center}
\begin{tabular}{lr}
\toprule
Reformulations per gold & 5 \\[0.1cm]
Accepted submissions & 10403 \\[0.1cm]
Rejected submissions & 550 \\[0.1cm]
Workers & 181 \\[0.1cm]
Avg \# submissions & 57.5  \\[0.1cm]
Work time per submission & 120.9s \\[0.1cm]
%Total cost & \$624 \\
\bottomrule
\end{tabular}
\caption{\label{tab:mturk}Crowd-sourced data collection effort.}
\end{center}
\end{table}

Table \ref{tab:rewrites}, shows an example of a crowd-sourced reformulation in this extended dialogue corpus.

\begin{table}[!h]
\scriptsize
\centering
\begin{tabular}{ll}
\toprule
Basis & What is the address []$_{zero}$? \\
\midrule
Gold & What is the address of the gas station Chevron? \\
\midrule
Crowd & (1) What is the address of the Chevron gas station? \\
& (2) Tell me the address for the Chevron gas station please. \\
& (3) What is the address to Chevron gas station? \\
& (4) Give me the address for Chevron gas station. \\
& (5) What is the address of the Chevron gas station? \\
\bottomrule
\end{tabular}
\caption{\label{tab:rewrites}Example rewrites with basis utterance, gold rewrites as well as 5 crowd-sourced rewrites.}
\end{table}

We also assessed each crowd-sourced rewrite quantitatively by determining F1 and BLEU~\cite{papineni2002bleu} score between the gold reformulation and each rewrite. To do this, 
we de-lexicalize slot values, using labels from the original corpus; we provide an example in Table \ref{tab:delexicalize} that results in F1 of 1.0 (all slot values are the same, meaning 
semantic similarity is high) with a BLEU score of 0.30 (low n-gram similarity indicative of syntactic/lexical variations).

\begin{table*}[!h]
%\scriptsize
\centering
\begin{tabular}{l l}
\toprule
Original & give me the address. \\
\midrule
Gold & give me the address of the mall Ravenswood 5 miles away. \\
\midrule
\midrule
Gold slots & give me the address of the [poi\_type0] [poi0] [distance0] \\
\midrule
Crowd slots &  give me the address to [poi0] [poi\_type0] that is [distance0] \\
\bottomrule
\end{tabular}
\caption{\label{tab:delexicalize} Delexicalizing slot values is one way to measure similarity between gold reformulations and crowd-sourced rewrites, used to determine F1 and Bleu 
similarity.}
\end{table*}

For the entire extended corpus of rewrites, we arrived at the similarity metrics presented in Table \ref{tab:similarity}. In addition to F1 and BLEU scores, we also counted how many 
slots each gold reformulation has on average compared to rewrites, where we see, for example, that the gold reformulations have on average almost one more slot value per utterance 
than rewrites; this is indicative of how unnatural it may be to compose utterances that specify an entity so unambiguously (e.g., ``Take me to the gas station Chevron two miles away.'')

\begin{table}[!h]
\centering
%\small
\begin{tabular}{lr}
\toprule
BLEU & 0.419 \\
F1 & 0.890 \\
Mean \# slots in each gold & 4.03 \\
Mean \# slots in each rewrite & 3.20 \\
Difference \# slots (gold vs rewrites) & 0.823 \\
\bottomrule
\end{tabular}
\caption{\label{tab:similarity}Similarity measures for gold reformulations vs crowd-sourced rewrites.}
\end{table}

\begin{figure}[ht!]
	\centering
	\begin{subfigure}{0.5\textwidth} % width of left subfigure
		\includegraphics[width=\textwidth]{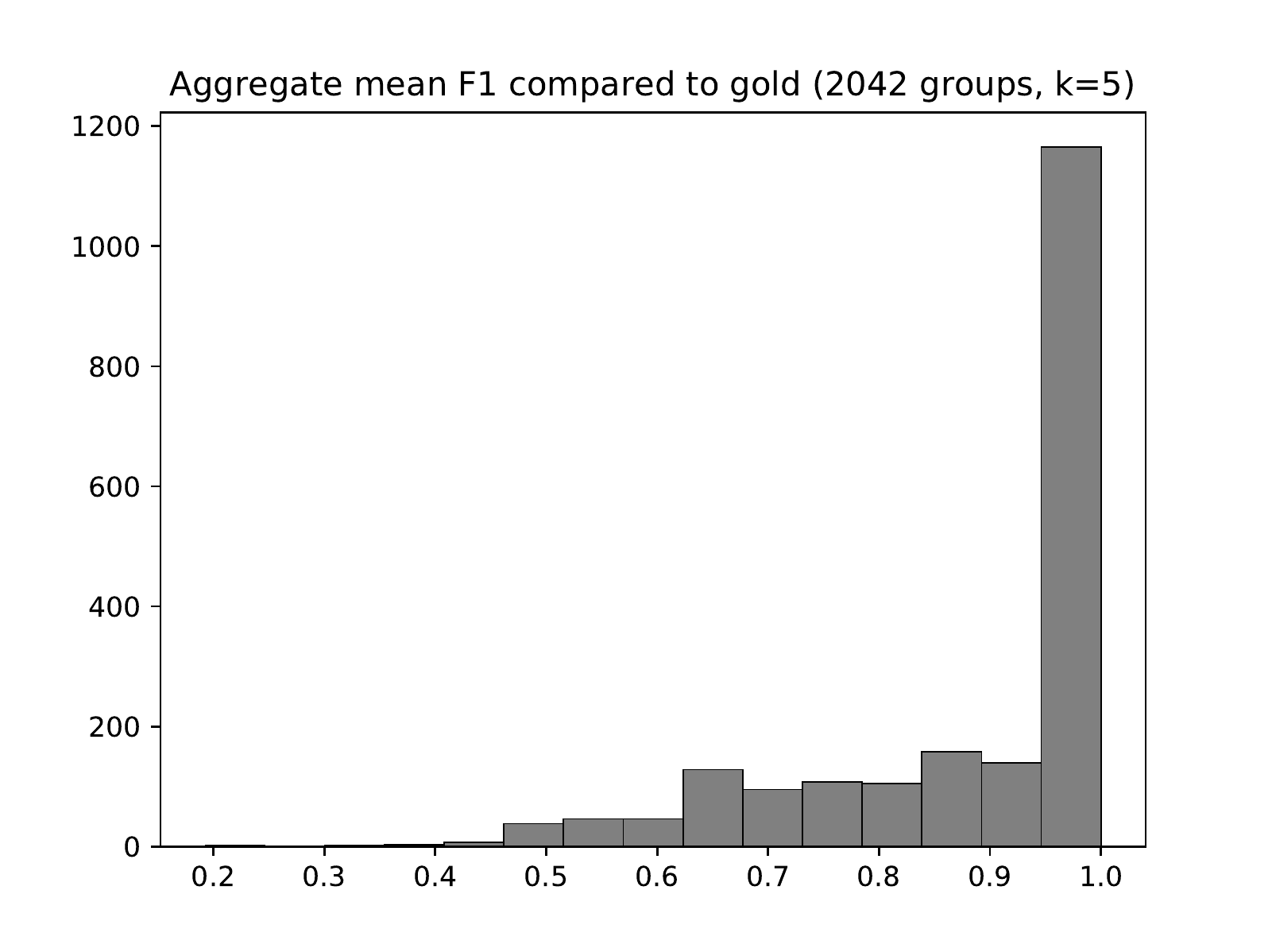}
		\caption{\label{fig:meanf1} Mean F1 scores for groups of crowd-sourced rewrites} 
		 % subcaption
	\end{subfigure}
	\vspace{1em} % here you can insert horizontal or vertical space
	\begin{subfigure}{0.5\textwidth} % width of right subfigure
		\includegraphics[width=\textwidth]{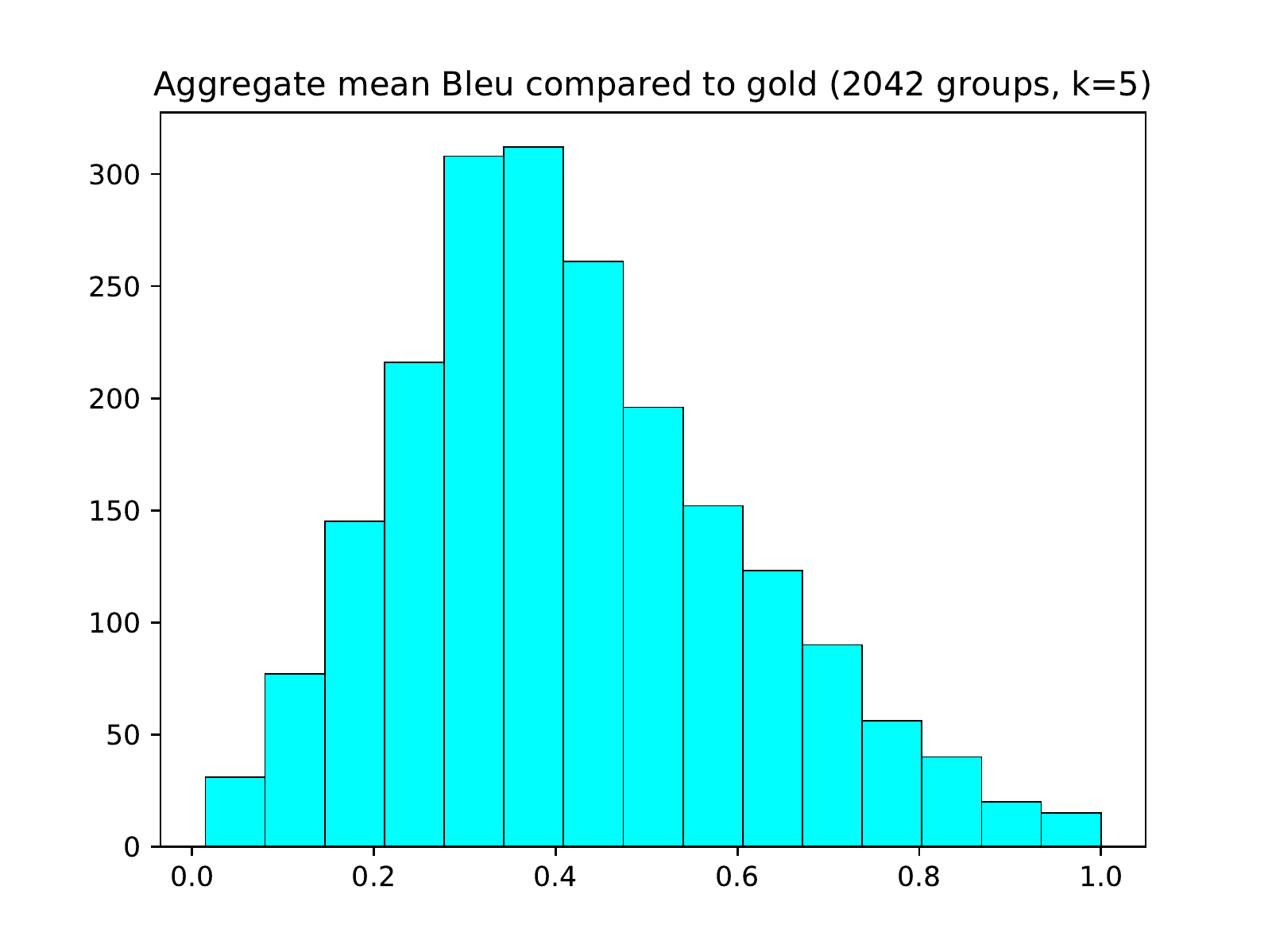}
		\caption{\label{fig:manebleu} Mean BLEU scores for groups of crowd-sourced rewrites} % subcaption
	\end{subfigure}
	\caption{\label{fig:hist1} Mean F1 and BLEU scores for groups of crowd-sourced rewrites} % caption for whole figure
\end{figure}

We also compared the five rewrites as a group to their corresponding gold reformulation: first, we grouped the rewrites for each dialogue (2042 total); next, we determined mean F1, 
BLEU, and difference in number of slots for each group pairwise compared to the gold reformulation; then, each group's F1 and BLEU scores were binned as shown in Figure 
\ref{fig:hist1}, where mean F1 scores are indicative of high in-group semantic similarity; and low BLEU scores indicative of syntactic and lexical variation within each group.

%\begin{figure}[!h]
%\centering
%\begin{minipage}{.25\textwidth}
%  \centering
%  \includegraphics[width=1.0\linewidth]{images/mturkers_mean_f1.eps}
%\end{minipage}%
%\begin{minipage}{.25\textwidth}
%  \centering
%  \includegraphics[width=1.0\linewidth]{images/mturkers_mean_bleu.eps}
%\end{minipage}
%\captionof{figure}{\label{fig:histf1}On the left, F1 scores for groups of crowd-sourced rewrites, with BLEU on the right.}
%\end{figure}

\section{Experiments and Analysis}
\label{sec:experiments}

\subsection{Establishing the baseline}

Spoken language understanding consists usually of two tasks, domain classification/intent determination (e.g., Weather, Navigation, etc.) and slot-filling which identifies the spans of 
text in the utterance assigned to a slot value-attribute pair given the domain. In the dialogue corpus used here , the three  domains and their corresponding slot values are: Weather 
(location, date, weather attribute); Navigation (point of interest type, point of interest, address, traffic information, distance); and Calendar scheduling (date, time, location, party, 
agenda). As a baseline, we first compare the original dialogues and the gold reformulations on the domain classification and slot-filling tasks. 

To assess this, we use a joint classifier for both tasks, an attention-based RNN for domain classification \& slot filling \citep{Liu2016, Mesnil2015}. We evaluate performance on two different inputs from the proposed dataset:
\begin{description}
\item[Original] The input is the original dialogue (concatenated user plus system utterances) and the output is the relevant slot values and domain at the user turn.
\item[Gold CQR] The input is the gold reformulation for the above user turn and the output is the relevant slot values and domain i.e we treat the dialogue as a single shot utterance.
\end{description}

For pre-processing, we encode the data using BIO tags. We perform the classification task on the two datasets and then compare the accuracy of the semantic labeler on the slots that both setups share (i.e., ignoring how the classifier does on the newly added slots in the gold reformulations). 

\begin{table}[h!]

\begin{center}
\begin{tabular}{|c|c|c|c|c|}
\hline
Input Type & Domain Classification  & Slot F1\\
& Accuracy & \\\hline\hline
Original & 0.98 & 0.89 \\\hline
\textbf{Gold CQR} & 0.98 & 0.91 \\\hline
\end{tabular}
\end{center}
\caption{Comparing original dialogue vs gold reformulations (Gold CQR) on two tasks domain classification and slot-filling. Domain classification accuracy between the original dataset and the reformulated query is similar. Slot-filling accuracy for reformulated query is higher than that for the original dataset.}
\label{tab:domain}
\end{table}%

Results in Table~\ref{tab:domain} indicate that while for domain classification there is no significant gain when comparing gold reformulations against the original dataset, for slot-filling task, increased prediction accuracy is evident for the gold reformulations. This suggests that query reformulation could lead to potentially better language understanding performance 
downstream as it is relatively easier to train and optimize NLU systems for single-shot utterances compared to multi-turn utterances.

\subsection{Query Rewriting Experiments}

In a second set of experiments, we would like to test  the query rewriting more directly. As described in Section~\ref{ssec:cqr}, we view this as summarizing a dialogue into a single 
utterance unambiguously specifying user intent. For the 
experiment, we delexicalize slot values using the canonical entity type from the original corpus (e.g., poi\_type = place of interest type), giving an example in Table \ref{tab:input}.   \\

{\renewcommand{\arraystretch}{1.2}
\begin{table}[!h]
\small
\begin{tabular}{l l}
\toprule
(input) & USER i need directions to a poi\_type0 \\
{} & SYSTEM i have a poi\_type0 that is distance0 \\
{} & USER give me the address \\
(output) & give me the address of poi\_type0 distance0 \\
\bottomrule
\end{tabular}
\caption{\label{tab:input} Input/output for CQR model}
\end{table}
}

\begin{figure}[ht!]
	\centering
	\begin{subfigure}{0.5\textwidth} % width of left subfigure
		\includegraphics[width=\textwidth]{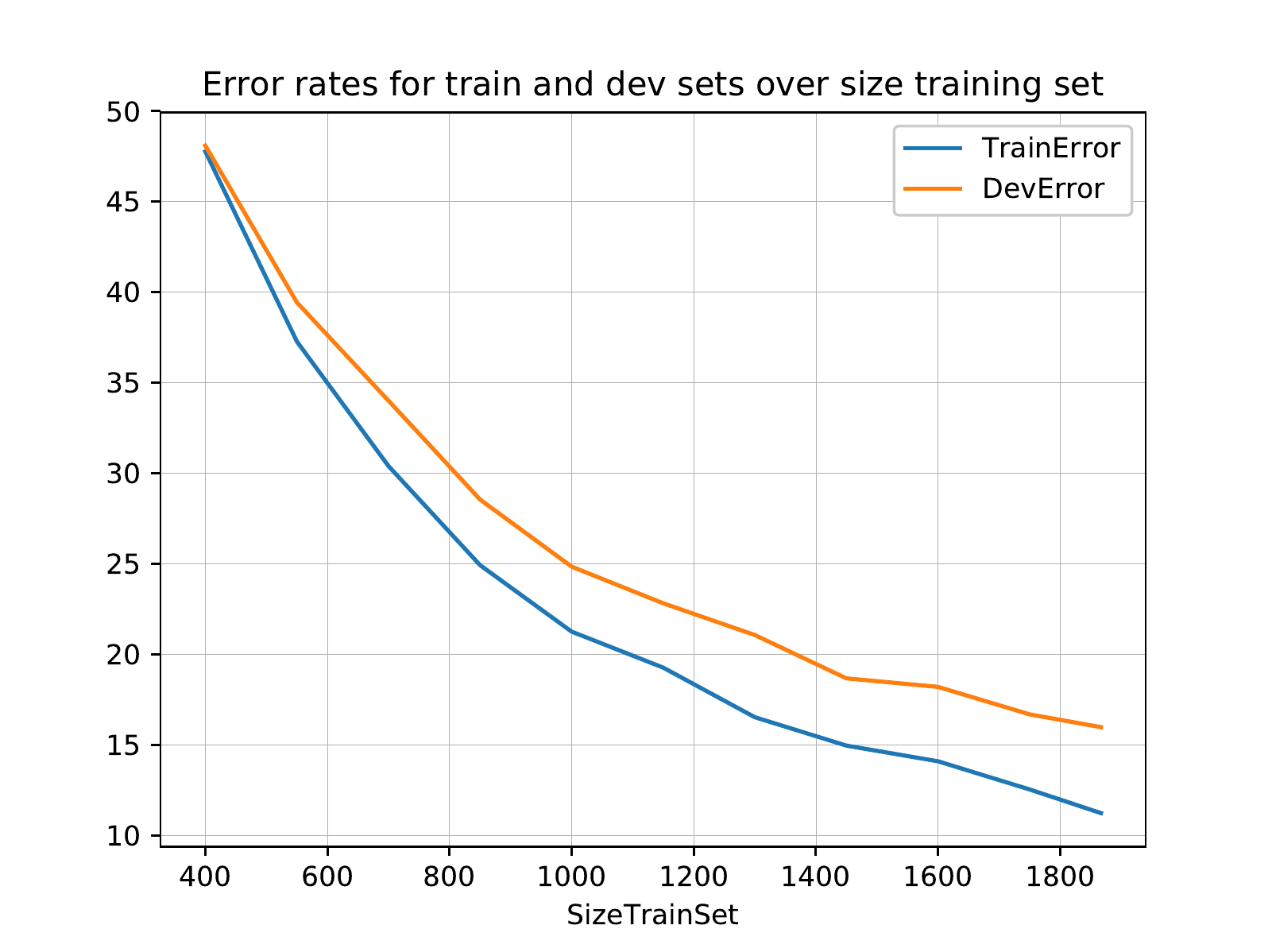}
		\caption{\label{fig:gold_error} Training error for gold rewrites} 
		 % subcaption
	\end{subfigure}
	\vspace{1em} % here you can insert horizontal or vertical space
	\begin{subfigure}{0.5\textwidth} % width of right subfigure
		\includegraphics[width=\textwidth]{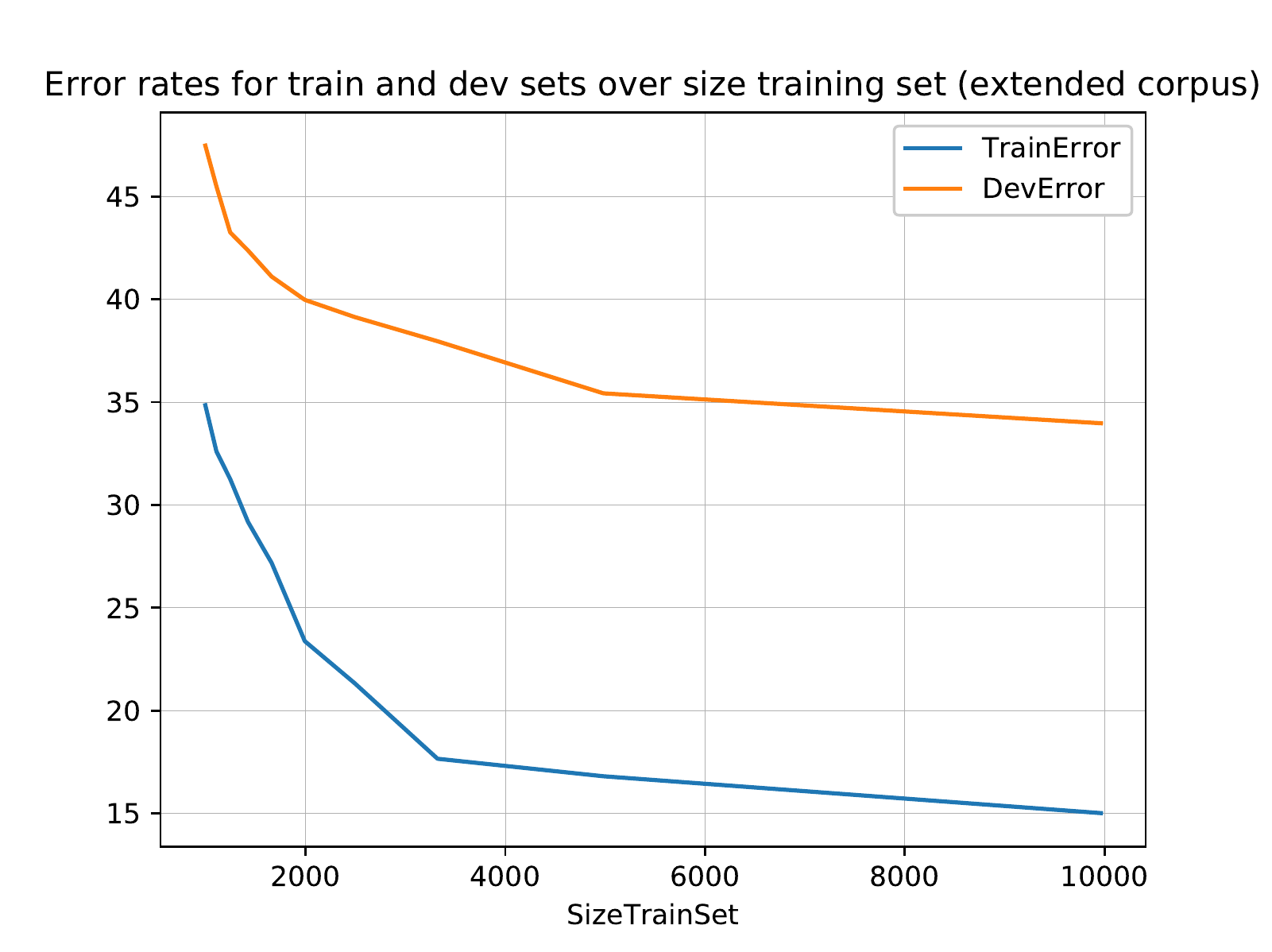}
		\caption{\label{fig:mturk_error} Training error for crowd-sourced Mturk rewrites} % subcaption
	\end{subfigure}
	\caption{\label{fig:error} Comparing gold reformulations vs Mturk extensions} % caption for whole figure
\end{figure}

We train two separate models drawing from different distributions: one of only gold reformulation data and the other including crowd-sourced reformulations. To quantify task 
complexity, we note that over the entire corpus, 67\% of slots are carried over from dialogue to reformulation, an indication of the non-triviality of the task. For reproducibility, we use the 
open source neural sequence modeling system OpenNMT~\cite{Klein2017}. The only hyper-parameter changed from initial settings is to remove {\it copy\_loss\_by\_seqlength}, which 
improves overall accuracy. Error rates for the training and dev sets are presented in the Figure \ref{fig:error} with observed accuracy metrics presented in Table \ref{tab:metrics}. The dev set error for the Mturk extensions is higher indicating that there is lexical diversity in the reformulations, which is also seen in the BLEU scores in  Table \ref{tab:metrics}. The entity F1 scores is higher, as the model has seen many more variations around the carryover entities. 

%\begin{figure}
%\centering
%\begin{minipage}{.25\textwidth}
%  \centering
%  \includegraphics[width=1.0\linewidth]{images/trainingerroroversize-eps-converted-to.pdf}
%\end{minipage}%
%\begin{minipage}{.25\textwidth}
%  \centering
%  \includegraphics[width=1.0\linewidth]{images/trainingerroroversizecomplete.eps}
%\end{minipage}
%\captionof{figure}{\label{fig:error}On the left, the error rate for gold reformulations; on the right, for the MTurk extension}
%\end{figure}

\begin{table}[!h]
\small
%\scriptsize
\centering
\begin{tabular}{lrrr}
\toprule
{} & train/dev/test & F1 & Bleu \\
\midrule
Gold & 2100/230/230 & 0.838 & 0.711 \\
Gold + MTurk & 10045/1279/1279 & \textbf{0.897} & 0.397 \\
\bottomrule
\end{tabular}
\caption{\label{tab:metrics}Accuracy metrics for contextual query rewrite task}
\end{table}

\section{Conclusion}

We show that anaphora is quite common in a single, human-created corpus \cite{Eric2017} of multi-turn dialogues used to train task-oriented, spoken dialogue systems. We  introduce 
contextual query rewrite (CQR), where the referring expressions resolution task is defined as query reformulations given the dialogue history. We show a principled approach to 
creating a corpus of query reformulations, and how this can be extended via crowd-sourcing. Two experiments demonstrate that query reformulations can be used to train high 
accuracy models for the task of generating fully unambiguous single-shot utterances as well as for more standard tasks of domain classification and slot filling, indicating that this 
approach may be suitable for anaphora resolution at larger scales. 

In future work, we intend to extend query reformulation for multiple languages, as well as to assess if anaphora resolution using query 
reformulation is also possible for longer dialogues. As a step towards improving dialogue systems in general and encouraging work on anaphora resolution specifically, we make our 
corpus publicly available.

\bibliography{cqr_dataset}
\bibliographystyle{acl_natbib}

%\appendix

%\section{Supplemental Material}
%\label{sec:supplemental}

\end{document}